\documentclass[journal,twoside,web]{ieeecolor}
\usepackage{tmi}
\usepackage{cite}
\usepackage{amsmath,amssymb,amsfonts}
\usepackage{xcolor}
\usepackage{multirow}
\usepackage{graphicx}
\usepackage{algorithm, algpseudocode}
\usepackage{float}
\usepackage{booktabs}   
\usepackage{pifont}  
\floatname{algorithm}{Algorithm}
\usepackage{amsmath, amssymb, amsfonts}
\usepackage{soul}
\usepackage{float}
\usepackage{caption}
\usepackage{subcaption}
\makeatletter
\newcommand*{\inlineequation}[2][]{%
  \begingroup
    \refstepcounter{equation}%
    \ifx\\#1\\%
    \else
      \label{#1}%
    \fi
    \relpenalty=10000 %
    \binoppenalty=10000 %
    \ensuremath{%
      #2%
    }%
    \hfill \@eqnnum
  \endgroup
}
\makeatother
\newcommand{\cmark}{\ding{51}}
\newcommand{\xmark}{\ding{55}}
\usepackage{graphicx,verbatim}

\usepackage{textcomp}
\def\BibTeX{{\rm B\kern-.05em{\sc i\kern-.025em b}\kern-.08em
    T\kern-.1667em\lower.7ex\hbox{E}\kern-.125emX}}
\begin{document}
\title{FeTTL: Federated Template and Task Learning for Multi-Institutional Medical Imaging}

\author{Abhijeet Parida, Antonia Alomar, Zhifan Jiang, Pooneh Roshanitabrizi, Austin Tapp,  \\Ziyue Xu, \IEEEmembership{Senior Member, IEEE}, Syed Muhammad Anwar, \IEEEmembership{Senior Member, IEEE}, \\Maria J. Ledesma-Carbayo, \IEEEmembership{Senior Member, IEEE}, Holger R. Roth,\\ Marius George Linguraru, \IEEEmembership{Senior Member, IEEE}
\thanks{Abhijeet Parida, Zhifan Jiang, Pooneh Roshanitabrizi, and Austin Tapp are with the Sheikh Zayed Institute for Pediatric Surgical 
Innovation, Children’s National Hospital, Washington, DC, USA. Abhijeet Parida is also affiliated with Universidad Polit\'{e}cnica de Madrid, Madrid, Spain (corresponding author’s e-mail: pabhijeet@childrensnational.org; other emails: zjiang@childrensnational.org; proshnani2@childrensnational.org; atapp@chidrensnational.org).\\ Antonia Alomar is with the Universitat Pompeu Fabra, Barcelona, Spain (e-mail: antonia.alomar@upf.edu).\\ Maria J. Ledesma-Carbayo is with the Universidad Polit\'{e}cnica de Madrid and CIBER-BBN, ISCIII, Madrid, Spain (e-mail: mj.ledesma@upm.es).\\Ziyue Xu and Holger R. Roth are with NVIDIA Corporation, Santa Clara, CA, USA (emails: ziyuex@nvidia.com; hroth@nvidia.com).\\Syed Muhammad Anwar and Marius George Linguraru are with the Sheikh Zayed Institute for Pediatric Surgical Innovation, Children’s National Hospital, Washington, DC, USA and the Departments of Radiology and  Pediatrics, George Washington University, Washington, DC, USA (e-mail: sanwar@childrensnational.org; mlingura@childrensnational.org).}}

\maketitle

\begin{abstract}
Federated learning enables collaborative model training across geographically distributed medical centers while preserving data privacy. However, domain shifts and heterogeneity in data often lead to a degradation in model performance. Medical imaging applications are particularly affected by variations in acquisition protocols, scanner types, and patient populations. To address these issues, we introduce Federated Template and Task Learning (\textit{FeTTL}), a novel framework designed to harmonize multi-institutional medical imaging data in federated environments. \textit{FeTTL} learns a global template together with a task model to align data distributions among clients. We evaluated \textit{FeTTL} on two challenging and diverse multi-institutional medical imaging tasks: retinal fundus optical disc segmentation and histopathological metastasis classification. Experimental results show that \textit{FeTTL} significantly outperforms the state-of-the-art federated learning baselines (\textit{p}-values $\mathbf{\le 0.002}$) for optical disc segmentation and classification of metastases from multi-institutional data. Our experiments further highlight the importance of jointly learning the template and the task. These findings suggest that \textit{FeTTL} offers a principled and extensible solution for mitigating distribution shifts in federated learning, supporting robust model deployment in real-world, multi-institutional environments.
\end{abstract}


\section{Introduction}

Machine learning (ML) improves modern healthcare systems in multiple ways, including disease diagnosis~\cite{ahsan2022machine}, treatment planning~\cite{dubey2023using}, and patient monitoring~\cite{ramirez2024machine}. These successes suggest that robust ML design could enable the deployment of scalable clinical decision support systems across institutions. However, data heterogeneity between sites impedes the generalization of models~\cite{rajendran2024learning}. Specifically, for medical imaging, variations in acquisition protocols and scanner hardware introduce substantial data distribution shifts between institutions~\cite{seoni2024all,parida2024quantitative}. Consequently, a model trained on data from a single site often performs poorly when applied to new and unseen data with non‑overlapping distributions. A natural way to improve generalization is to aggregate data across institutions for models to be exposed to a richer diversity of data. However, privacy and ethical restrictions often prevent data sharing, motivating federated learning (FL) approaches as a privacy-preserving alternative.

\subsection{Data Sharing and Federated Learning}

Aggregating imaging data from multiple sites exposes ML models to a wider range of image characteristics, thus improving robustness~\cite{rajendran2024learning,parida2024quantitative}.  Privacy regulations such as the Health Insurance Portability and Accountability Act (HIPPA) in the United States and the General Data Protection Regulation (GDPR) in the European Union, together with ethical considerations, can restrict the sharing of patient data by institutions \cite{rahman2024addressing} and hamper the training of models with diverse data. This fundamental challenge can be resolved using FL by enabling distributed training that preserves privacy. In FL, institutions keep their data locally and share only model parameters with a central aggregator. \textit{FedAvg}~\cite{fedavg} is a common FL technique that averages each of the local client models on the central aggregator server. However, \textit{FedAvg} assumes that the data in each client are independent and identically distributed (IID), an unrealistic expectation in many medical applications. 

Numerous FL variants like \textit{FedBN}~\cite{li2021fedbn}, \textit{FedProx}~\cite{li2020federated}, \textit{FedDG}~\cite{feddg} and \textit{FedHarmony}~\cite{fed48} have been proposed to address non‑IID data arising from client‑specific scanners and demographic changes~\cite{fedXX}. \textit{FedBN}~\cite{li2021fedbn}  uses client-specific batch normalization to adapt to feature changes. This approach preserves local batch norm statistics at each client, but prevents the server from shipping a single model to all clients, causes large performance drops when batch norm layers are not trained~\cite{wang2025population}, and is not readily adaptable to architectures that use other types of normalization layers, such as vision transformers. 

\textit{FedProx}~\cite{li2020federated} adds a proximal term that regularizes local models toward the server to reduce drift. This approach is sensitive to parameters\cite{Yu2025-ca}) and computationally expensive due to the additional computation of the proximal gradients.

\textit{FedDG}~\cite{feddg} collects external amplitude spectra from a fast Fourier transform (FFT) on the server. These amplitudes are combined with the local phase on the client side to create stylized images. The system is then trained on the stylized image using episodic meta-learning to learn features that work across different domains. However, FFT-based amplitude alignment may not capture protocol changes, label conventions, or structural variations.

\textit{FedHarmony}~\cite{fed48} builds on \textit{FedProx} by aggregating client‑specific feature statistics through a shared encoder. This approach approximates each client's latent representation as a Gaussian distribution and operates on the assumption that the data distribution under shift is close to a global Gaussian. 

Although FL mitigates privacy concerns, performance limitations occur when deviations from the training data distributions are significant. Therefore, it is prudent to consider the data perspective as a means of minimizing data deviation during FL training.

\subsection{Style Learning for Domain Generalization}
Methods such as \textit{CycleGAN}~\cite{cyclegan2017,sandfort2019data}, \textit{StarGAN}~\cite{choi2018stargan,choi2020stargan}, and Adaptive Instance Normalization (\textit{AdaIN})~\cite{huang2017arbitrary} view robustness across data sources as a data harmonization problem rather than model fine-tuning. Other approaches, such as data style learning, aim to improve generalization to new domains by transforming the input data instead of altering the model itself~\cite{dataalchemy,iwasawa2021test}.

\textit{CycleGAN}~\cite{cyclegan2017,sandfort2019data} is one of the first methods designed to translate images between two domains by coupling an adversarial loss function with a cycle consistency constraint\cite{cyclegan2017}. Being an adversarial method, \textit{CycleGAN} faces mode collapses and trains separate models for every pair of sites, which does not scale well when more than two acquisition protocols are involved.

\textit{StarGAN}~\cite{choi2018stargan,choi2020stargan} overcomes the scaling problem of \textit{CycleGAN} with a single conditional generator and discriminator for multi‑domain translation by using a one-shot domain label to select the style. \textit{StarGAN} also presumes that domain labels are discrete and known. However, in practice, medical images often exhibit continuous scanner variability that is not well captured by categorical tags.

To avoid the need for discrete domain labels, the class of \textit{AdaIN}  harmonizers~\cite{huang2017arbitrary}~\cite{oh2022cxr} align the channel-wise mean and variance of content features with those of a sampled style code. This allows real-time, arbitrary style transfer without requiring adversarial training. Compared to GAN-based methods, \textit{AdaIN} is computationally light and scales gracefully to many application domains. However, with \textit{AdaIN} style harmonization, synthetic styles are sampled from a parametric Gaussian, leading to a higher likelihood of excluding rare but clinically relevant appearances.

\textit{DataAlchemy}~\cite{dataalchemy} tackles site changes by performing data calibration to make them IID, which avoids the need to increase training data by image augmentation. A whitening–coloring transform, learned from in-source images, harmonizes client-specific protocols without touching model weights~\cite{dataalchemy}. The calibration is confined to color/intensity shifts and may not correct geometric shifts or other artifacts.

These style-based methods highlight the promise of harmonization, but typically require access to raw images or domain labels, making them difficult to embed within federated training. This motivates a closer look at how existing approaches complement or conflict with each other.

\subsection{Summary of Prior Approaches and Open Gaps}
Heterogeneity in acquisition protocols, hardware, and annotation practice produces
pronounced distribution shifts across imaging sites, undermining the reliability of convolutional and transformer models trained at a single institution.
In summary, existing FL strategies preserve privacy
but struggle with non‑IID data, while style learning techniques
harmonize appearances but require access to raw pixels or site labels and do not address collaborative training at multiple sites.
\begin{table}[h]
  \centering
  \caption{Comparison of representative methods.
           \cmark\;=\;satisfies criterion, \xmark\;=\;does not.}
  \label{tab:rw_summary}
\centering
\scriptsize
\resizebox{\columnwidth}{!}{%

 \begin{tabular}{lccc}
    \toprule
    \textbf{Method} & \textbf{\begin{tabular}[c]{@{}c@{}}Federated \\ Learning\end{tabular}} & \textbf{\begin{tabular}[c]{@{}c@{}}Style \\ Harmonization\end{tabular}} &\textbf{\begin{tabular}[c]{@{}c@{}}Joint Harmonization  \& \\ Task Learning\end{tabular}} \\ \hline

CycleGAN\cite{cyclegan2017} & \xmark & \cmark & \xmark\\
StarGAN\cite{choi2018stargan} & \xmark & \cmark & \xmark\\
AdaIN Harmonizer\cite{huang2017arbitrary,oh2022cxr} & \xmark & \cmark & \xmark\\
DataAlchemy\cite{dataalchemy} & \xmark & \cmark & \xmark\\ 
FedAvg\cite{fedavg} & \cmark & \xmark & \xmark\\
FedProx\cite{li2020federated} & \cmark & \xmark & \xmark\\
FedBN\cite{li2021fedbn} & \cmark & \xmark & \xmark\\

FedHarmony\cite{fed48} & \cmark & \xmark & \xmark \\
FedDG\cite{feddg} & \cmark & \cmark  & \xmark\\
\hline
\textbf{\textit{FeTTL} (ours)} & \cmark & \cmark & \cmark \\
    \bottomrule
  \end{tabular}
}
\end{table}

Among previous works, as shown in Table~\ref{tab:rw_summary}, only \textit{FedDG} attempts to combine federated training with appearance harmonization. However, its reliance on FFT amplitude alignment limits its ability to capture protocol-level changes, annotation conventions, or structural variations between institutions.  This reveals an important scientific gap: there is no harmonization framework that (1) preserves strict patient-level privacy, (2) harmonizes cross-site styles, and (3) jointly optimizes harmonization with downstream task performance.

\begin{figure*}[ht]
    \centering
    \includegraphics[width=\textwidth]{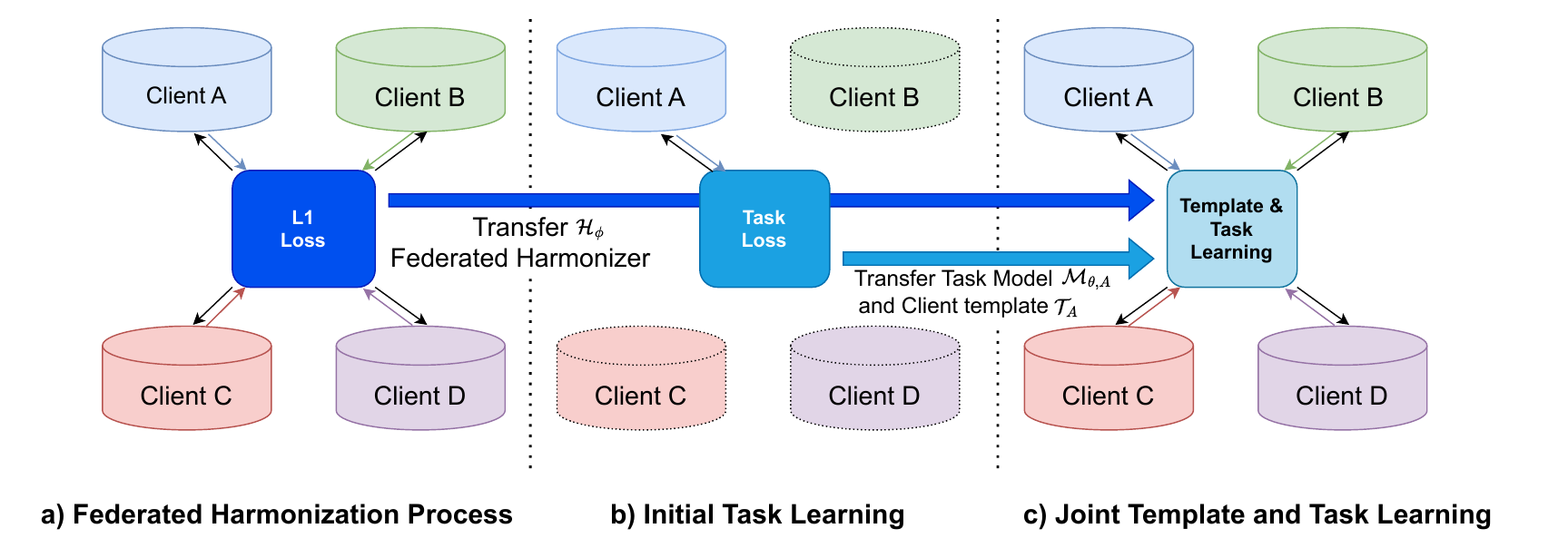}
    \caption{\textbf{Federated Template and Task Learning (\textit{FeTTL}) framework} showing the major steps of a) federated harmonization process, b) initial task learning, and c) the joint template and task learning. The arrows show the flow of information between the client and the server. Global model data is sent to each client shown in black arrows, while the colored arrows represent local data collection.}
    \label{fig:fetll-framework}
\end{figure*}

To bridge this gap, we propose a novel framework that
(1) learns a global image template through a federated whitening–coloring transform and
(2) optimizes the harmonization and the downstream task objectives in a single loop. This framework is expected to establish a practical path towards robust medical AI methods that preserve privacy in multi-site setting.

\subsection{Our Contributions}
We introduce \textbf{Federated Template and Task Learning (\textit{FeTTL})}, a novel FL framework to harmonize non‑IID medical images when multiple clients collaborate. \textit{FeTTL} integrates image harmonization and task learning into a seamless process to train a shared model without ever sharing raw patient data. Our primary contributions are summarized as follows.

First, we propose a federated approach to training a style-template-based harmonization model based on \textit{DataAlchemy}, which was originally designed for centralized datasets. This FL approach uses the whitening–coloring transform and aggregates the resulting statistics, enabling image‑level harmonization without sharing raw pixel data. Aggregating these statistics allows the model to align appearance across non‑IID clients while keeping anatomical structures intact.

Second, we introduce a novel method called \textit{FeTTL}, a federated single‑loop optimization process that jointly learns a global image template for all clients, the harmonization model and a downstream task model.

Finally, we evaluate \textit{FeTTL} on two challenging and diverse medical imaging tasks: retinal fundus disc segmentation using five geographically and clinically diverse datasets, and histopathological metastases classification using the CAMELYON16 dataset of two different clients. In both tasks, \textit{FeTTL} significantly improves downstream performance over other FL methods that are built to handle non-IID data (see Section \ref{sec:results}).

Collectively, these contributions introduce a practical and scalable solution for harmonizing federated medical imaging data with the potential to improve model generalization between clients and enabling clinical institutions with limited data resources to participate in collaborative modeling without compromising privacy.

\section{Methods}

The proposed \textit{FeTTL} framework consists of three main stages, as illustrated in Fig.~\ref{fig:fetll-framework}: 
(1) federated harmonization to train a harmonization model $\mathcal{H}_\phi$ across all sites, 
(2) initial task learning at each site to train a site-specific task model $\mathcal{M}_{\theta,k}$ and extract a style template $\mathcal{T}_k$, and
(3) joint template and task learning to collaboratively learn a global task model $\mathcal{M}_{\theta,\text{server}}$ and global style template $\mathcal{T}_{\text{server}}$.

Let $I \in \mathbb{R}^{H \times W \times 3}$ be a 2D medical image, where $H, W$, and $3$ are the height, width, and number of channels, respectively, at a site $S_k \in \{S_1, S_2, \dots\}$ in an FL setting.

\noindent \textbf{Task model} $\mathcal{M}_{\theta,k}$ is a network with parameters $\theta$ trained on data from the site $S_k$. It maps an input image $I$ to a task-specific output $y$. 

\noindent \textbf{Template} $\mathcal{T}$ is the encoded representation of the style characteristics, obtained from a source image $I$. 

\noindent \textbf{Harmonization model} $\mathcal{H}_\phi$ is a federated harmonization network with an encoder $enc(\cdot)$ and a decoder $dec_\phi(\cdot)$ with trainable weights $\phi$. The harmonized image $I_h$ was computed as:
\begin{equation}
    I_h = \mathcal{H}_\phi(I_0, \mathcal{T}_s), \quad \text{where } \mathcal{T}_s \leftarrow enc(I_s).
    \label{eqn:harm}
\end{equation}

In the following subsections, we take a closer look at each of the three stages of \textit{FeTTL}.

\subsection{Federated Harmonization Process}

Our harmonization process is based on the whitening and coloring transforms of previous work in style transfer~\cite{li2017universal,dataalchemy}, adapted to the federated setting to reduce inter-site variability in style while preserving anatomical content. The whitening and coloring transforms are chosen for their differentiability property compared to non adversarial style transfer methods.

\subsubsection{Training}

Let $enc(\cdot)$ be a pre-trained encoder with frozen weights, and $dec_\phi(\cdot)$ a trainable reconstruction decoder. At each client site $S_k$, we trained $dec_\phi$ to minimize $L_1$ reconstruction loss for local images $I \in S_k$:
\begin{equation}
\mathcal{L}_{rec} = \| I - dec_\phi(enc(I)) \|_1,
\end{equation}

The choice of norm $\| \cdot \|_1$ over $\| \cdot \|_2$, for $\mathcal{L}_{rec}$ is made to encourage less blurry reconstruction and to make the model robust to outliers in the data. For each client, $dec_\phi$ is trained locally. After training, the updated weights are sent to the central server. The server aggregated client models using \textit{FedAvg}~\cite{fedavg}, as shown in Fig. \ref{fig:fetll-framework}a. This decentralized training process is repeated for multiple communication rounds to obtain a global $dec_\phi(\cdot)$. The encoder $enc(\cdot)$ is kept fixed across sites to ensure a consistent representation of features.

\subsubsection{Inference}
For harmonization, inference was carried out in two stages: \textit{whitening} and \textit{coloring}. The style correlations of an input image, $I_0$, are removed using the whitening transform~\cite{li2017universal}:
\begin{equation}
f_0 = E_0 D_0^{-\frac{1}{2}} E_0^\top enc(I_0),
\end{equation}

\noindent where $E_0 D_0 E_0^\top$ is the eigen-decomposition of the covariance matrix of the centered $enc(I_0)$ and $f_0$ are features with structural information of $I_0$.  Then, the coloring transform is applied using a template $\mathcal{T}_s$. The template $\mathcal{T}_s = enc(I_s)$ is derived from a reference image $I_s$ representing the required style. The harmonized feature $f_h$ is obtained using

\begin{equation}
f_h = E_s D_s^{\frac{1}{2}} E_s^\top f_0,
\end{equation}

\noindent where $E_s D_s E_s^\top$ is the covariance matrix of template $\mathcal{T}_s$. Finally, the harmonized image $I_h$ is reconstructed using

\begin{equation}
I_h = dec_\phi(f_h) = \mathcal{H}_\phi(I_0, \mathcal{T}_s).
\end{equation}

\subsection{Initial Task Learning}

After training $\mathcal{H}_\phi$, we performed the initial task learning as shown in Fig.~\ref{fig:fetll-framework} b. We selected the largest site $S_k$ to initialize the task model and the template. The task model $\mathcal{M}_{\theta,k}$ is trained on an image $I \in S_k$ to minimize task-specific loss
\begin{equation}
\mathcal{L}_{\text{task}} = \mathbb{E}[\ell(I, y_{gt})],
\end{equation}
where $y_{gt}$ denotes the ground-truth label and $\ell(\cdot,\cdot)$ is the loss function appropriate to the task (e.g. cross entropy for classification, Dice loss for segmentation).

After training, template $\mathcal{T}_k \leftarrow enc(I_s)$ is extracted from a randomly sampled image $I_s \in S_k$ using the harmonization encoder $enc(\cdot)$. The resulting $\mathcal{M}_{\theta,k}$ and $\mathcal{T}_k$ served as initializations for the global task model $\mathcal{M}_{\theta,\text{server}}$ and the global template $\mathcal{T}_{\text{server}}$ in the FL phase.

\subsection{Joint Template and Task Learning}

Once $\mathcal{H}_\phi$ and the task model $\mathcal{M}_{\theta, k}$ are trained, we use an FL framework that jointly optimizes the global template and the downstream task model between sites (Fig.~\ref{fig:fetll-framework}c). The learning is performed in two steps. First, image-style representations are aligned, and the task network is simultaneously adapted to improve performance on non-IID client data. Second, we aggregate client updates on the task model and templates.  The general procedure is summarized in Algorithm~\ref{algo:1}.

\begin{algorithm}[t]
\caption{Federated Template and Task Learning
(\textit{FeTTL})}
\begin{algorithmic}[0]
\small
\State \textbf{1. Federated Harmonization}: Train $\mathcal{H}_\phi$ using FedAvg 
\State \textbf{2. Initial Task Learning} at  site $S_k$:
\State Train task model $\mathcal{M}_{\theta,k}$, extract $\mathcal{T}_{\text{k}} \leftarrow enc(I_s)$; $Is \in S_k$
\State \textbf{3. Joint Template and Task Learning}:
\State \textit{Initialize server template and task model} :  \State $\mathcal{T}_{\text{server}} \leftarrow \mathcal{T}_{\text{k}}; \mathcal{M}_{\theta,k,\text{server}}\leftarrow \mathcal{M}_{\theta,k}$
\State
\For{round in $R$ }
\For{each site $S_i$}
    \State \textit{Initialize client template and task model} : 
    \State$\mathcal{T}^{(i)}_{\text{client}} \leftarrow \mathcal{T}_{\text{server}}$
    \State$\mathcal{M}^{(i)}_{\theta,k,\text{client}} \leftarrow \mathcal{M}_{\theta,k,\text{server}}$
    \State 
    \State \textit{Perform Optimization at $S_i$}: 
    \For{$(I^{(i)}, y_{gt}^{(i)})$ in $S_i$}
    \State
    \State \textit{Compute local loss $\mathcal{L}$}:
    \State \begin{equation}
    \begin{aligned}
    &\mathcal{L} \leftarrow \sum\mathbb{E}  \ell(\mathcal{M}_{\theta,k}(\mathcal{H}_\phi(I^{(i)},\mathcal{T}^{(i)}_{\text{client}})), y_{gt}^{(i)}) \\
    &=
    \sum\mathbb{E} [\ell(y^{(i)}, y_{gt}^{(i)}) ]
    \end{aligned}
        \label{eqn:loss}
    \end{equation}
    \State
    \State \textit{Update parameters using gradient descent:}
    \State \begin{equation}
     \begin{aligned}[t]
&\mathcal{T}^{(i)}_{\text{client}} \leftarrow \mathcal{T}^{(i)}_{\text{client}} - \eta \nabla_{\mathcal{T}^{(i)}_{\text{client}}} \mathcal{L} \\
&\mathcal{M}_{\theta,k, client}^{(i)} \leftarrow \mathcal{M}_{\theta,k, client}^{(i)} - \beta \nabla_{\mathcal{M}_{\theta,k, client}^{(i)}} \mathcal{L}
\end{aligned}
        \label{eqn:update}
    \end{equation}
\EndFor
\EndFor
\State \textit{Global Aggregation}: 
\State \begin{equation}
     \begin{aligned}[t]
&\mathcal{T}_{\text{server}} \leftarrow \frac{1}{n} \sum_{i=1}^{n} \mathcal{T}^{(i)}_{\text{client}} \\
&\mathcal{M}_{\theta,k,\text{server}} \leftarrow \frac{1}{n} \sum_{i=1}^{n} \mathcal{M}_{\theta,k, client}^{(i)}
\end{aligned}
        \label{eqn:agg}
    \end{equation}

\EndFor
\end{algorithmic}
\label{algo:1}

\end{algorithm}

\subsubsection{Local Training}
At the client site $S_k$, the server-provided template $\mathcal{T}_{\text{server}}$ and the task model $\mathcal{M}_{\theta,k , \text{server}}$ are initialized as local copies $\mathcal{T}_{\text{client}}^{(i)}$ and $\mathcal{M}_{\theta,k, \text{client}}^{(i)}$. Then, each input image $I^{(i)}$ is harmonized using $\mathcal{H}_\phi$ and the local template, as $ I_h^{(i)} = \mathcal{H}_\phi(I^{(i)}, \mathcal{T}_{\text{client}}^{(i)})$.

The prediction $y^{(i)} = \mathcal{M}_{\theta,k,\text{client}}^{(i)}(I_h^{(i)})$ is compared with the ground truth $y_{gt}^{(i)}$ using a task-specific loss function $\ell(\cdot,\cdot)$.
Although initially derived from a reference image using
$enc(\cdot)$, the template
along with the task model are treated as differentiable parameters and updated using gradients from the task-specific loss.

\subsubsection{Global Aggregation}
Once local optimization is completed, the server aggregates the client updates to obtain the global parameters. These updated global parameters are then redistributed to clients for the next communication round. This iterative process is designed to allow \textit{FeTTL} to adaptively align site-specific styles while improving task performance among all clients.

\section{Experimental Setup}
\subsection{Materials}

We evaluated the proposed \textit{FeTTL} framework on two downstream tasks: (1) optical disc segmentation of retinal fundus images and (2) classification of metastases in histopathological images of the lymph node section. Table~\ref{tab:dataset} summarizes the datasets used and the respective training/validation/test splits.

\begin{table}[ht]
\caption{\textbf{Summary of datasets}: splits, sites for FL, and the number of images used for all the experiments.}
\centering
\resizebox{\columnwidth}{!}{%
\begin{tabular}{cccccc}
\toprule
\textbf{Task} & \textbf{Dataset} &  & \textbf{Train} & \textbf{Validation} & \textbf{Test} \\ \hline
\multirow{5}{*}{\textbf{\begin{tabular}[c]{@{}c@{}}Optical\\ disc\\ segmentation\end{tabular}}} & \begin{tabular}[c]{@{}c@{}}Site A\\ Drishti-GS1\cite{drishti}\end{tabular} &  & 50 & 25 & 26 \\
 & \begin{tabular}[c]{@{}c@{}}Site B\\ RIGA-BinRushed4\cite{riga}\end{tabular} &  & 98 & 49 & 48 \\
 & \begin{tabular}[c]{@{}c@{}}Site C\\ RIGA-Magrabia\cite{riga}\end{tabular} &  & 47 & 24 & 23 \\
 & \begin{tabular}[c]{@{}c@{}}Site D\\ RIGA-Messidor\cite{riga}\end{tabular} &  & 230 & 115 & 115 \\
 & \begin{tabular}[c]{@{}c@{}}Site E\\ REFUGE\cite{refuge}\end{tabular} &  & 400 & 200 & 200 \\ \hline
\multirow{4}{*}{\textbf{\begin{tabular}[c]{@{}c@{}}Histopathological\\ metastases\\ classification\end{tabular}}} & \multirow{2}{*}{\begin{tabular}[c]{@{}c@{}}Site X\\ CAMELYON 16\cite{bejnordi2017diagnostic}\end{tabular}} & healthy & 77,204 & 33,306 & 20,183 \\
 &  & metastases & 52,000 & 58,000 & 34,000 \\ 
 & \multirow{2}{*}{\begin{tabular}[c]{@{}c@{}}Site Y\\ CAMELYON 16\cite{bejnordi2017diagnostic}\end{tabular}} & healthy & 71,447 & 11,551 & 8,449 \\
 &  & metastases & 60,000 & 8,000 & 12,000 \\ \bottomrule
\end{tabular}
}
\label{tab:dataset}
\end{table}

\noindent
\textbf{1) Retinal fundus optical disc segmentation:} We used three public datasets that span five sites: Site A of Drishti-GS1~\cite{drishti}, Sites B, C, and D from RIGA~\cite{riga}, and Site E from REFUGE~\cite{refuge}, as in~\cite{jiang2023fair}. All data sets include color fundus images annotated by expert clinicians for optical disc segmentation ($y_{gt}$, ground truth). Drishti-GS1 was collected at Aravind Eye Hospital (India) with subjects aged 40 to 80 years, including normal and glaucomatous eyes.  The three sites in RIGA come from MESSIDOR in France, Bin Rushed and Magrabi in Saudi Arabia. The sources differ in cameras, illumination, and clinical focus (e.g., DR screening in MESSIDOR). REFUGE contains images acquired in China with two cameras (Zeiss Visucam 500 at 2124×2056 and Canon CR-2 at 1634×1634), including both normal and glaucomatous eyes. Each image was resized to $256\times256$ pixels and normalized to $[0,1]$. No additional pre-processing was done, such as cropping or contrast enhancement.

\noindent
\textbf{2) Histopathological metastases classification:} We used the CAMELYON16 dataset~\cite{bejnordi2017diagnostic} from two institutions: Site X (Radboud University Medical Center, Nijmegen with Pannoramic 250 Flash II scanner; 3DHISTECH) and Site Y (University Medical Center, Utrecht with NanoZoomer-XR Digital slide scanner C12000-01; Hamamatsu Photonics). The data set consists of 400 whole slide images of sections of lymph nodes stained with hematoxylin and eosin reagents from patients with breast cancer. Metastases and healthy patches of size $256\times256$ were extracted using the coordinates provided by Baidu Research~\cite{ncrf}, which indicate the center of the regions with metastases or healthy tissue.  All patches were normalized to $[0,1]$. To prevent leakage of patient data between training, validation, and test sets, we used the whole-slide splits provided by \cite{dataalchemy}, consisting of 314 training patients, 42 validation patients, and 44 testing patients split across 2 sites.

These data sets were selected to represent various acquisition protocols, imaging devices, and clinical characteristics often encountered in multi-institutional settings. For example, Drishti-GS1 and MESSIDOR differ in their field of view and intended clinical use. Similarly, the CAMELYON16 sites vary in staining procedures and scanning hardware. Additionally, there are notable variations in sample volume between sites (e.g., Site D has 230 training samples while Site C has only 47), reflecting common data imbalance scenarios. Each site was treated as an independent client in our FL setup, and the non-IID characteristics of the data provide a realistic benchmark for FL in medical imaging.

\subsection{Model Training}

All models were trained using the AdamW~\cite{loshchilov2017decoupled} optimizer with default settings from PyTorch. Standard data augmentations were applied: random color jitter and brightness jitter within $\pm 0.2$, horizontal and vertical flips with 50\% probability, random rotations of $\pm 5^\circ$ and random 90-degree rotations with 50\% probability. 

\subsubsection{Federated harmonizer}

Following the setup in~\cite{dataalchemy}, we used a VGG-19~\cite{vgg} architecture for both segmentation and classification harmonization. The encoder $enc(\cdot)$ included layers up to \texttt{conv\_3\_3} of VGG-19 and the model weights were kept frozen. The decoder $dec_\phi(\cdot)$ was initialized as a mirrored architecture and trained using FedAvg~\cite{fedavg} on a FL simulator. Each client optimized $dec_\phi$ with a batch size of 96 and a learning rate of $1e^{-4}$. After each FL round, updated decoder weights were sent to the server for aggregation. This process was repeated for 100 FL rounds.

\begin{figure*}[ht]
\centering
\includegraphics[width=\textwidth]{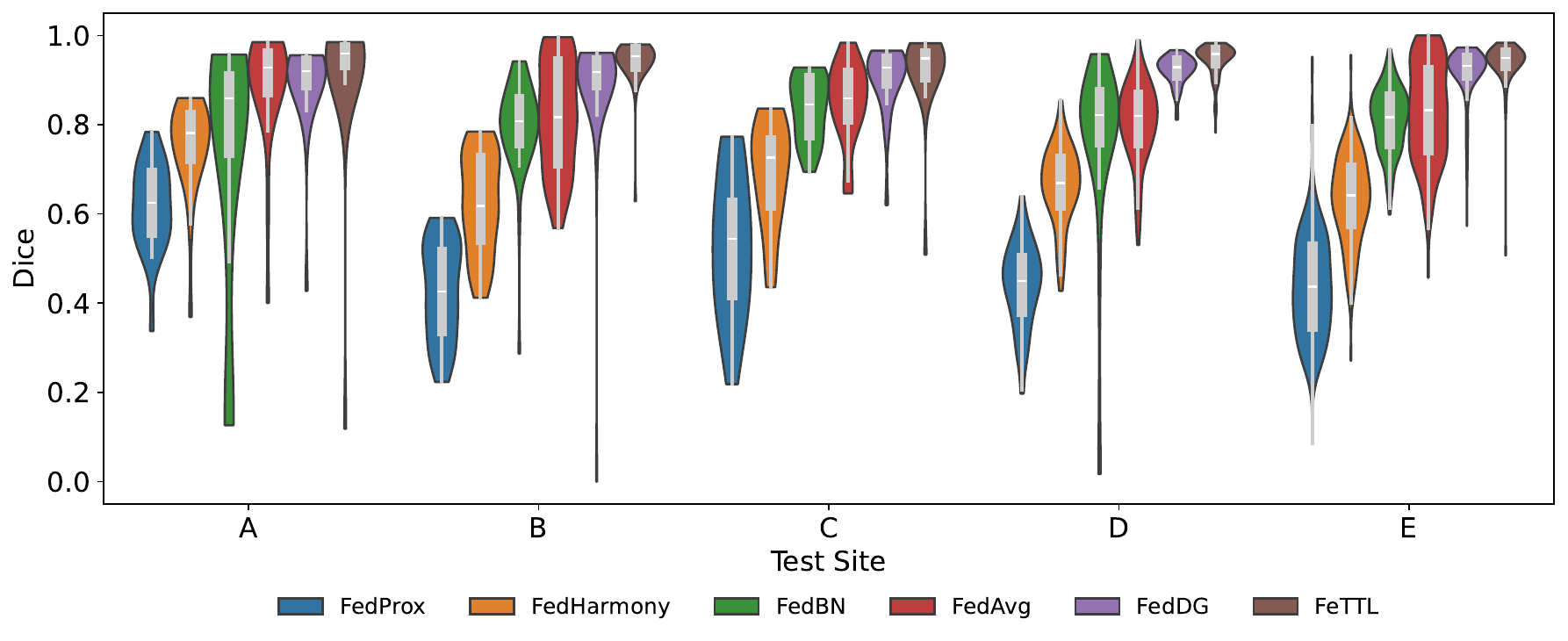}
\caption{\textbf{Qualitative performance of \textit{FeTTL}} versus \textit{FedBN}, \textit{FedProx}, \textit{FedDG} and \textit{FedHarmony} for retina disc segmentation. The violin plots show the distribution of test Dice scores and the box plot on the $y$-axis for a test site  on the $x$-axis. The box blot shows the median and the inter quartile range. }
\label{fig:comp-violin}
\end{figure*}

\subsubsection{Initial Task Model}

For the segmentation task, we used a U-Net~\cite{unet} with instance normalization, as implemented by default in MONAI~\cite{MONAI}. The initial model $\mathcal{M}_{\theta,k}$ was trained on Site E for 5 epochs using a learning rate of $5e^{-4}$ and a batch size of 96, optimized using the Dice loss. For the classification task, a ResNet-34 architecture in \cite{ncrf} was trained on Site Y for 5 epochs with a learning rate of $1e^{-4}$ and a batch size of 256 using cross-entropy loss. The sites selected for this phase were the largest among all available sites for each task. 

\subsubsection{Template and Task Model}\label{sec:trg-task-model}
The joint learning of the segmentation task and template was carried out in all sites {A, B, C, D, E}. Local optimization was performed at each site for $\mathcal{M}_{\theta,k, \text{client}}$ and $\mathcal{T}_{\text{client}}$ for 5,000 iterations, batch size of 128, and learning rate $1e^{-4}$ for 10 FL rounds. For the classification task, local optimizations for {X, Y} was done for 10 epochs with a learning rate $5e^{-5}$ for 10 FL rounds. The model with the best validation scores was used for testing in all sites. 

\subsubsection{Other Federated Baselines}
We compared our methods with other FL baselines, i.e., \textit{FedBN}, \textit{FedProx}, \textit{FedDG} and \textit{FedHarmony}  trained with the same FL setup in Section \ref{sec:trg-task-model}. Since \textit{FedBN} requires batch norm layers, we modified the U-Net architecture to use them instead of the instance norm. \textit{FedDG} implementation for histopathological metasteses classification assumes the logits of the ResNet as the embeddings for meta learning.  

\subsubsection{Implementation Environment}
All experiments were implemented in PyTorch v2.7.1 with MONAI v1.5.0~\cite{MONAI} and NVFlare v2.4.2rc3~\cite{nvflare} for FL simulator experiments. Training was carried out on a machine with NVIDIA RTX A5000 GPUs (24GB VRAM) and 64GB RAM. Random seeds 64 bits long were fixed in runs for reproducibility. All clients participated in each round, and sequential client updates were simulated on a single GPU.

\subsection{Evaluation and Statistical Analysis}

Model performance was assessed using task-specific metrics. For the segmentation task, we used the Dice similarity coefficient between predicted and ground-truth masks. Statistical significance was tested using the Wilcoxon signed-rank test with a significance threshold $p$-value $< 0.05$ by combining all predictions in the five sites.

For the classification task, we used the Area Under the Precision–Recall Curve (AUPR), which is appropriate for class-imbalanced data. We applied the DeLong test to compare the significance between receiver operating characteristics, also using a $p$-value $< 0.05$ by combining all predictions at both sites.

For qualitative visualization of \textit{FeTTL}'s harmonization effect, we use t-SNE scatter plots~\cite{tsne}. t-SNE was used to reduce the dimension of images in the HSV space to 2D. 

\section{Results}\label{sec:results}
\subsection{Retinal Image Segmentation}\label{sec:results-comparision}

\begin{table}[ht]
\caption{\textbf{Comparison of \textit{FeTTL} with other FL baselines} for retinal optic disc segmentation across five diverse clinical sites (A–E). Test Dice scores are reported per site, along with the overall average across all sites.   \textit{p}-values are calculated using Wilcoxon signed-rank test by combining all predictions in the five sites. \textit{FeTTL} significantly outperformed all baselines, demonstrating generalization under non-IID data settings.}
\centering
\resizebox{0.99\columnwidth}{!}{%
\begin{tabular}{l|cccccc|c}
\toprule
 & \multicolumn{6}{c|}{\textbf{Testing Site}} & \multirow{2}{*}{\textbf{\textit{p}-value}} \\
 & \textbf{A} & \textbf{B} & \textbf{C} & \textbf{D} & \textbf{E} & \textbf{Avg} &  \\ \midrule
Centralized & 0.909& 0.925 & 0.899 & 0.927 & \textbf{0.943} & 0.921 & 0.023 \\ \midrule
FedProx\cite{li2020federated} & 0.616 & 0.425 & 0.530 & 0.442 & 0.440 & \multicolumn{1}{l|}{0.491} & $<0.001$ \\
FedBN\cite{li2021fedbn} & 0.733 & 0.799 & 0.837 & 0.779 & 0.809 & 0.791 &  $<0.001$\\
FedHarmony\cite{fed48}& 0.765 & 0.615 & 0.711 & 0.679 & 0.648 & 0.684 & $<0.001$ \\
FedAvg\cite{fedavg} & 0.894 & 0.826 & 0.876 & 0.816 & 0.839 & 0.850 &  $<0.001$\\
FedDG\cite{feddg} & 0.890 & 0.892 & 0.890 & 0.920 & 0.915 & 0.901 &  0.002\\
\midrule
FeTTL (ours) & \textbf{0.918} & \textbf{0.948} & \textbf{0.934} & \textbf{0.944} & 0.941 & \textbf{0.937} &  - \\
\bottomrule
\end{tabular}
}
\label{tab:retina-seg}
\end{table}
First, we investigated the performance of our proposed \textit{FeTTL} framework against various FL baselines on the retinal fundus optical disc segmentation task, i.e., \textit{FedDG}, \textit{FedAvg}, \textit{FedBN}, \textit{FedProx}, and \textit{FedHarmony}. Table~\ref{tab:retina-seg}  and Fig. \ref{fig:comp-violin} summarize the performance results of the Dice similarity coefficient on the test set of the five clinical sites (A–E) under non-IID conditions. Statistical significance was tested using the Wilcoxon signed-rank test with a significance threshold $p$-value $< 0.05$ by combining all predictions in the five sites.

\textit{FeTTL} achieved the highest average Dice score of 0.937, significantly outperforming all federated baselines ($p$-value $\le 0.002$ for all, see Table ~\ref{tab:retina-seg}). Furthermore, centralized training, which assumes data pooling--an impractical scenario in privacy-sensitive clinical settings--also had a significantly lower performance of Dice score of 0.921 ($p$-value = 0.023). This improvement in performance highlights \textit{FeTTL}’s ability to improve inter-client performance by template learning and task optimization.

\begin{table*}[!ht]
\caption{\textbf{Ablation study of \textit{FeTTL} on retinal disc segmentation.} Performance (Dice score) is reported across testing sites A–E when progressively adding harmonization, template learning, and task learning components. The proposed \textit{FeTTL} achieves the highest average Dice, significantly outperforming all baselines.}
\centering
\resizebox{\textwidth}{!}{%
\begin{tabular}{l|cc|cc|c||cccccc}
\toprule
 & \multicolumn{2}{c|}{\textbf{Harmonization}} & \multicolumn{2}{c|}{\textbf{Template Learning}} & 
 \textbf{Task} & \multicolumn{6}{c}{\textbf{Evaluation Site}}  \\
 & At client & Federated & At client & Federated & \textbf{Learning} &\textbf{A} & \textbf{B} & \textbf{C} & \textbf{D} & \textbf{E} & \textbf{Avg} \\ \midrule
AdaIN Harmonizer & \cmark &  &  &    &  & 0.897 & 0.840 & 0.877 & 0.842 & 0.837 & 0.859 \\
DataAlchemy & \cmark &  & \cmark &  &  &   0.889 & 0.864 & 0.892 & 0.883 & 0.891 & 0.884 \\
Federated DataAlchemy &  & \cmark & \cmark &  &  &   0.912 & 0.905 & 0.903 & 0.896 & 0.916 & 0.906 \\
Global Template Learning &  & \cmark &  & \cmark &  &   0.912 & 0.910 & 0.909 & 0.906 & 0.918 & 0.911 \\
\textit{FeTTL} (task model from scratch) &  & \cmark &  & \cmark & \cmark  & 0.785 & 0.892 & 0.803 & 0.875 & 0.900 & 0.851 \\
\textit{FeTTL} (ours) &  & \cmark &  & \cmark & \cmark &  \textbf{0.918} & \textbf{0.948} & \textbf{0.934} & \textbf{0.944} & \textbf{0.941} & \textbf{0.937}  \\
\bottomrule
 \end{tabular}
}
\label{tab:ablation}
\end{table*}
\begin{figure*}[h]
\centering
\includegraphics[width=\textwidth]{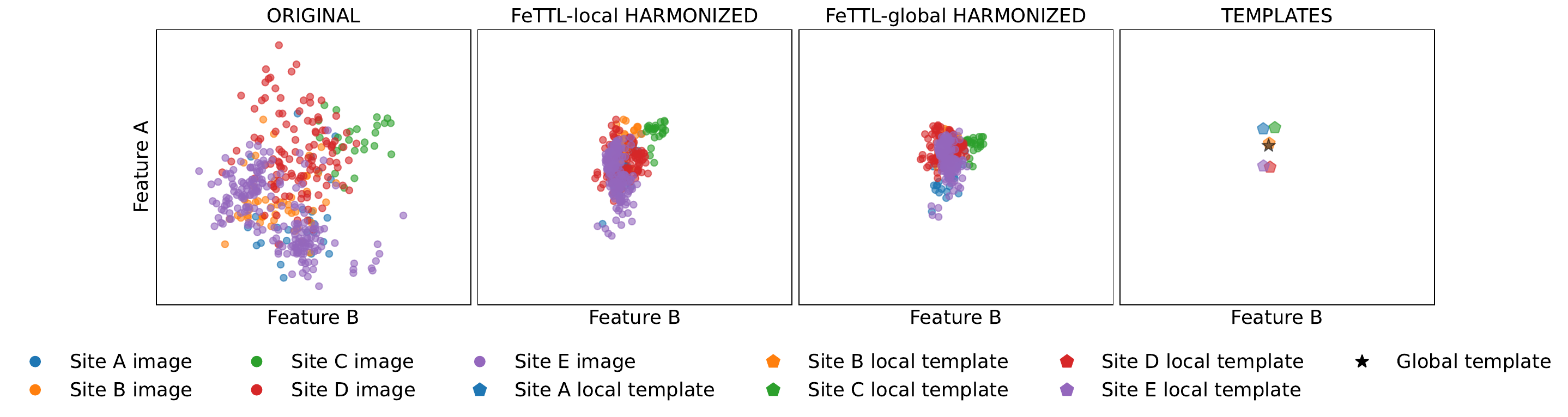}
\caption{\textbf{Clustering performance} of  \textit{FeTTL} on retinal images using t-SNE \cite{tsne} illustrates the representation of the test image on \textit{HSV} space for the original, the \textit{FeTTL-local} and \textit{FeTTL-global} and the templates obtained, respectively. Each color represents a different site. The cluster of harmonized images appear to be more compact for \textit{FeTTL-global} over \textit{FeTTL-local}.}
\label{fig:scatter-plot}
\end{figure*}

\subsection{Ablation Study}\label{sec:results-ablation}

We conducted an ablation study to systematically assess the contribution of harmonization, template learning, and task learning in both client-wise and federated settings. Table~\ref{tab:ablation} summarizes the results.

In Table~\ref{tab:ablation}, we see that the baseline \textit{AdaIN} which applies client-wise harmonization only achieved an average Dice of 0.859. Adding client-specific template learning using \textit{DataAlchemy}~\cite{dataalchemy} significantly improved Dice performance to 0.884 ($p$-value  = 0.03). Extending \textit{DataAlchemy} to Federated \textit{DataAlchemy} using a federated harmonizer further significantly improved the mean Dice to 0.906 ($p$-value  = 0.022). This demonstrates the benefit of sharing harmonization across clients while retaining local template learning.

Replacement of client-specific templates with a global template yielded an additional improvement of Dice from 0.906 to 0.911 ($p$-value  = 0.12). This suggests that global style templates may be more effective than site-specific templates.

Next, we incorporated task learning. Training the segmentation model from scratch under federated harmonization and global template learning resulted in unstable performance with a Dice score of 0.851. In contrast, our proposed \textit{FeTTL}, which jointly optimizes federated harmonization, global template learning, and downstream task learning, from an initial task model, achieved the best performance with an average Dice of 0.937. This represents a substantial improvement over both the client-wise and the federated baselines of the ablation ($p$-value $< 0.001$).

In general, these results highlight the benefits of federated harmonization, global template learning, and joint task optimization. Their joint contribution and integration in \textit{FeTTL} leads to generalized performance at non-IID evaluation sites.

\subsection{Impact of Global Template Learning on Harmonization}\label{sec:results-local-global}

Next, we investigated impact of \textit{FeTTL}'s global template initialization on disc segmentation. We defined two \textit{FeTTL} variants: (1) \textit{FeTTL-local}: during FL, no global aggregation of the template occurred, so each site learned a unique local template; (2) \textit{FeTTL-global}: local  templates were aggregated to a single global template for all sites.

\textit{FeTTL-global} average Dice score of 0.937 was a modest improvement over the  \textit{FeTTL-local} Dice score of 0.928 ($p$-value = 0.12). Figure \ref{fig:scatter-plot} shows the t-SNE plots~\cite{tsne}  for images at different sites. The plot also shows the learned templates from \textit{FeTTL-global} and the \textit{FeTTL-local} after projecting the templates to the \textit{HSV} image space. 
Furthermore, Fig.~\ref{fig:scatter-plot} shows that both \textit{FeTTL-local} and \textit{FeTTL-global} effectively harmonize client images, as the harmonized images form a single cluster.
\textit{FeTTL-global} appears to be more compact than \textit{FeTTL-local}. 

\subsection{Impact of Template Initialization}\label{sec:init}
\begin{figure*}[h]
\centering
\includegraphics[width=\textwidth]{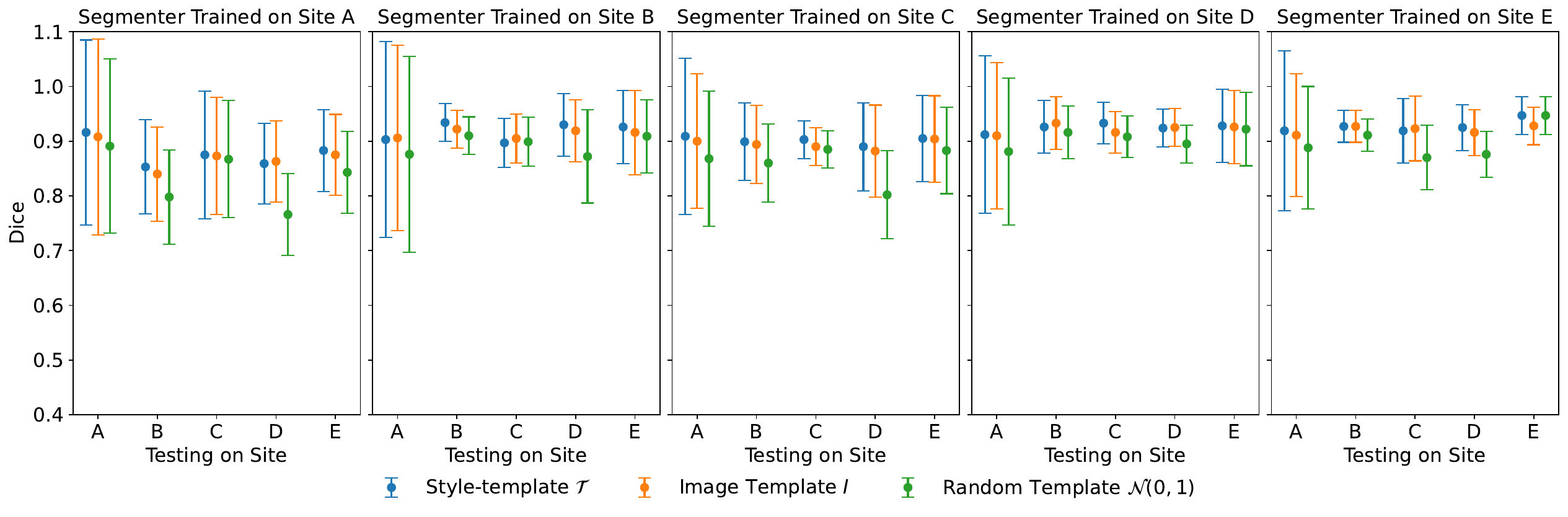}
\caption{\textbf{Impact of the template initialization} on \textit{FeTTL} for retinal disc segmentation. The figure shows whisker plots of $\mathcal{M}_{\theta,k}$ for different $\mathcal{T}_{\text{global}}$ initializations on the test data: (1) an image from site $k$, $\mathcal{I}_k$; (2) a template from $k$, such that $\mathcal{T}_k= enc(I_k)$; or (3) a random $\mathcal{T}_{\text{global}} \sim \mathcal{N}(0,1)$. The best performance on Dice scores was achieved for the \textit{FeTTL} using a template from $k$, such that $\mathcal{T}_k= enc(I_k)$.}
\label{fig:temp_init}
\end{figure*}
We also examined how the global template affected performance of \textit{FeTTL} in disc segmentation. We considered three $\mathcal{T}_{\text{global}}$: (1) an image from $k$, $I_k$; (2) a template from $k$, where $\mathcal{T}_k= enc(I_k)$; and (3) random noise $\sim \mathcal{N}(0,1)$.

Figure~\ref{fig:temp_init} shows the template initialization results for sites \textit{A}–\textit{E}. From Fig.~\ref{fig:temp_init}, we see that $\mathcal{T}_k = enc(\mathcal{I}_k)$ achieved comparable mean Dice scores with $\mathcal{T}_k = \mathcal{I}_k$ ($p$-values = 0.10, 0.10, 0.23, 0.47, and 0.09). In contrast, it can be seen that random noise initialization $\mathcal{T}_k \sim \mathcal{N}(0,1)$ resulted in significantly lower Dice scores ($p$-values $<$ 0.01, $<$ 0.01, 0.02, 0.08, and $<$ 0.01). These findings demonstrate that $\mathcal{T}_k = enc(\mathcal{I}_k)$  offers a generalizable starting point for \textit{FeTTL} across sites.

\subsection{Performance on Metastases Classification}\label{sec:results-histopathology}

In addition, we investigated the performance of our proposed \textit{FeTTL} framework against various FL baselines for the histopathological metastasis classification task. Table~\ref{tab:clf}  summarizes the results of the AUPR scores of the test set of the two distinct clinical sites (X and Y) under non-IID condition.

Among all federated baselines \textit{FeTTL} achieved the highest AUPR score of 0.681 for the combined test set of both sites, significantly outperforming the federated baselines of \textit{FedAvg}, \textit{FedBN}, \textit{FedProx}, and \textit{FedHarmony}\ ($p$-value $ \leq 0.002$ for all, see Table ~\ref{tab:clf}). The \textit{FeTTL} AUPR score of 0.681 is a moderate improvement over \textit{FedDG} AUPR score of 0.660 ($p$-value = 0.092).

\begin{table}[ht]
\caption{\textbf{Comparison of FeTTL with other FL baselines} for histopathological metastases classification. Test AUPR scores are reported per site, along with the AUPR of the combined test sites. \textit{p}-values are calcualted using DeLongs test combining all predictions at
both sites. \textit{FeTTL} outperformed all baselines, demonstrating generalization under non-IID data settings.}
\centering
\scriptsize
\resizebox{0.80\columnwidth}{!}{%
\begin{tabular}{l|ccc|c}
\toprule
 & \multicolumn{3}{c|}{\textbf{Testing site}} & \multirow{2}{*}{\textbf{\textit{p}-value}} \\
 & \textbf{X} & \textbf{Y} & \textbf{Both} &  \\ \midrule
Centralized & 0.888 & 0.902 & 0.887 & $<0.001$ \\ \midrule
FedProx~\cite{li2020federated} & 0.777 & 0.718 & \multicolumn{1}{l|}{0.634} & $<0.001$ \\
FedAvg~\cite{fedavg} & 0.761 & 0.704 & 0.638 & $<0.001$ \\
FedBN~\cite{li2021fedbn} & 0.666 & 0.512 & 0.603 &  $<0.001$\\
FedDG~\cite{feddg} & 0.799 & 0.607  & 0.660  & 0.092 \\
FedHarmony~\cite{fed48} & 0.762 & 0.700  & 0.640 & 0.002\\ \midrule
FeTTL(ours) & 0.793 & 0.722 & 0.681 & - \\
\bottomrule
\end{tabular}
}
\label{tab:clf}
\end{table}

\section{Discussion}

 Overall, \textit{FeTTL} is designed to address the challenges posed by non-IID medical imaging data originating at multiple clinical sites. Our joint training approach is particularly advantageous for sites with limited labeled data, as it enables them to benefit from the shared optimization of both the template and the task model. With our proposed approach, \textit{FeTTL} consistently outperformed the state-of-the-art FL baselines in various imaging tasks and performed similarly or exceeded centralized training performance. 

For example, as shown in Table~\ref{tab:retina-seg}, \textit{FedAvg} is unable to handle non-IID data effectively. \textit{FedBN} -- although designed to mitigate domain change through client-specific batch normalization statistics -- suffers from ambiguity in choosing inference statistics. In addition, both \textit{FedProx} and \textit{FedHarmony} rely on proximal terms to stabilize training, which introduce additional hyperparameters that are difficult to optimize and often degrade performance in practice. 

\textit{FeTTL} and \textit{FedDG} are the only methods in our comparison that employ site-specific harmonization strategies, which are key to handling non-IID variations. While \textit{FedDG} relies on FFT-based style harmonization, non-linear global template learning in \textit{FeTTL} approach captures better the inter-site style variability when aggregated over multiple sites. Hence, \textit{FeTTL} performed as well as or exceeded centralized analyses, most likely due to the increased capacity of the learnable template parameters in addition to the learnable task network. 

Our ablation studies (Table~\ref{tab:ablation}) illustrate the contribution of each component within \textit{FeTTL}. Harmonization alone, via \textit{AdaIN}, yielded a modest improvement over baseline \textit{FedAvg}. Integrating client-specific template learning through \textit{DataAlchemy} further increased performance, highlighting the importance of preserving site-specific statistical patterns. Extending template learning to the federated setting also boosted Dice scores, reflecting the benefits of aggregating style information between clients. Finally, jointly optimizing global template and downstream task models provided the largest gain, underscoring the synergistic effect of harmonization and task-specific adaptation in non-IID FL. 

The visualization in Fig.~\ref{fig:scatter-plot} highlights the impact of global template learning. \textit{FeTTL-local}, which learns only site-specific templates, performs comparable to \textit{FeTTL-global}, but the t-SNE clusters of harmonized images are visually more compact for \textit{FeTTL}-\textit{global}. This indicates that the shared global template reduces the variability between sites in image representations compared to using a site-specific template, and also facilitates more robust downstream task learning.

The selection of the starting template is critical, as shown in Fig.~\ref{fig:temp_init}. Initialization with a site-specific encoded template yields the best performance compared to initialization with raw images or random noise. These results indicate that the model requires a structured starting point for effective harmonization. In our framework, the initial task model and the site-specific template were derived from the largest participating site. This choice was motivated by the assumption that a site with more data and greater diversity would provide a stronger and more generalizable initialization for the downstream federated optimization. Importantly, the task model and template are both fine-tuned collaboratively during the FL phase, thus, site-specific biases in the initialization can be corrected. However, this heuristic selection could be optimized, particularly in cases where the largest site is not representative of the overall data distribution or contains domain-specific artifacts. Alternatively, the most representative site could be based on inter-site similarity metrics, ensemble pretraining, or meta-learning-based initialization schemes.

Therefore, \textit{FeTTL} is a generalizable technique for various modalities and medical imaging tasks, as reflected in the classification of histopathological metastases and the segmentation of retinal images. However, \textit{FeTTL} has certain limitations; (1) expanding \textit{FeTTL} to 3D magnetic resonance imaging remains a challenge due to the lack of robust 3D feature extractors similar to VGG~\cite{vgg}, (2) while \textit{FeTTL} outperformed centralized training in some cases, this advantage may be contingent on the learnable template representation; understanding the theoretical limits of such gains warrants future investigation.

In summary, \textit{FeTTL} presents three key takeaways for FL deployment in medical imaging applications. 
\begin{itemize}
  \item Integrating template and task learning into a single optimization loop improves model generalization, and harmonization should be embedded within the training process. 
\item Using a shared global style template leads to the alignment of feature distributions and consistent performance between clients. 
\item Employing lightweight harmonization methods over adversarial approaches facilitates robust, equitable, and privacy-preserving AI deployment in diverse clinical environments.
\end{itemize}

\section{Conclusion}
\textit{FeTTL} is a new FL framework that improves the analysis of non-IID medical images when multiple clients collaborate. \textit{FeTTL} integrates image harmonization, global template learning, and task learning into a federated process for every client to benefit from a shared model without sharing raw patient data. Our experiments on various medical imaging datasets and tasks demonstrate that \textit{FeTTL} outperforms baseline FL methods in improving model performance. To conclude, \textit{FeTTL} is a promising framework for federated medical imaging applications that is generalizable to different types of image data and adaptable to various image analysis tasks.

\bibliographystyle{IEEETran}
\bibliography{refs}

\end{document}